\documentclass{article}
\usepackage{spconf,amsmath,graphicx}
\usepackage{multirow}
\usepackage{ctable}
\usepackage{subfigure}

\definecolor{myred}{RGB}{192, 0, 0}
\definecolor{myblue}{RGB}{0, 112, 192}
\definecolor{mygreen}{RGB}{0, 176, 80}

\title{DOD-CNN: Doubly-injecting Object Information for Event Recognition}
\name{Hyungtae Lee$^{\star\dagger}$~~~~~~~~~~~~~~~~~~~~~~~~Sungmin Eum$^{\star\dagger}$~~~~~~~~~~~~~~~~~~~~~~~~Heesung Kwon$^{\star}$}
\address{$^{\star}$US Army Research Laboratory~~~~~~~~~~~~~~~~~~~~~~~~$^{\dagger}$Booz Allen Hamilton Inc.\thanks{Copyright 2019 IEEE. Published in the IEEE 2019 International Conference on Acoustics, Speech, and Signal Processing (ICASSP 2019), scheduled for 12-17 May, 2019, in Brighton, United Kingdom. Personal use of this material is permitted. However, permission to reprint/republish this material for advertising or promotional purposes or for creating new collective works for resale or redistribution to servers or lists, or to reuse any copyrighted component of this work in other works, must be obtained from the IEEE. Contact: Manager, Copyrights and Permissions / IEEE Service Center / 445 Hoes Lane / P.O. Box 1331 / Piscataway, NJ 08855-1331, USA. Telephone: + Intl. 908-562-3966.}
}

\begin{document}
%
\maketitle
\begin{abstract}
Recognizing an event in an image can be enhanced by detecting relevant objects in two ways: 1) indirectly utilizing object detection information within the unified architecture or 2) directly making use of the object detection output results. We introduce a novel approach, referred to as Doubly-injected Object Detection CNN (DOD-CNN), exploiting the object information in both ways for the task of event recognition. The structure of this network is inspired by the Integrated Object Detection CNN (IOD-CNN) where object information is indirectly exploited by the event recognition module through the shared portion of the network. In the DOD-CNN architecture, the intermediate object detection outputs are directly injected into the event recognition network while keeping the indirect sharing structure inherited from the IOD-CNN, thus being `doubly-injected'. We also introduce a batch pooling layer which constructs one representative feature map from multiple object hypotheses. We have demonstrated the effectiveness of injecting the object detection information in two different ways in the task of malicious event recognition.
\end{abstract}
\section{Introduction}
\label{sec:intro}
When figuring out what is happening in a photo, we typically make use of its relevant objects depicted in the scene. For example, if a bride, a bridegroom, and flowers are shown, we can naturally infer that it is highly likely to be related to a wedding event. Similarly, automatically recognizing such an event in an image should be empowered by exploiting relevant object information.


There are two approaches to exploit the object detection information in assisting the task of event recognition. The first approach is to make use of a separately constructed object detection module and its output for boosting the event recognition. In this approach, the object detection results can either be directly fed into the event recognition module ~\cite{TAlthoffACMMM2012,MJainCVPR2015,YChaoICCV2015} or be integrated with the event recognition output via a late fusion ~\cite{LWangCVPRW2015,LWangICCVW2015,RRobinsonIROS2015,HLeeIROS2016,HLeeDCS2016,YCaoICIP2017,YChaoWACV2018,HLeeICASSP2018}. The second approach is to transfer the object information by sharing the network weights between the object detection and event recognition and co-learning them in a unified architecture.

Eum et al.~\cite{SEumICIP2017} showed the effectiveness of the second approach by devising the IOD-CNN (Integrated Object Detection CNN) architecture which consists of three networks for event recognition, rigid object detection, and non-rigid object detection, where some initial layers are shared across all three tasks. Eum et al. claim that the enhancement in the event recognition performance was achieved by indirectly transferring the relevant object information by sharing certain portion of the weights in the network during the co-learning of multiple tasks. In \cite{SEumICIP2017}, we also note that applying a late fusion over three different tasks bring an additional performance increase. This observation provides another evidence that both approaches (direct and indirect usage of object information) are complementary to each other when pulling up the event recognition performance.

\begin{figure}[t]
\begin{minipage}[b]{1.0\linewidth}
  \centering
  \centerline{\includegraphics[width=\linewidth,trim=5mm 5mm 5mm 5mm,clip]{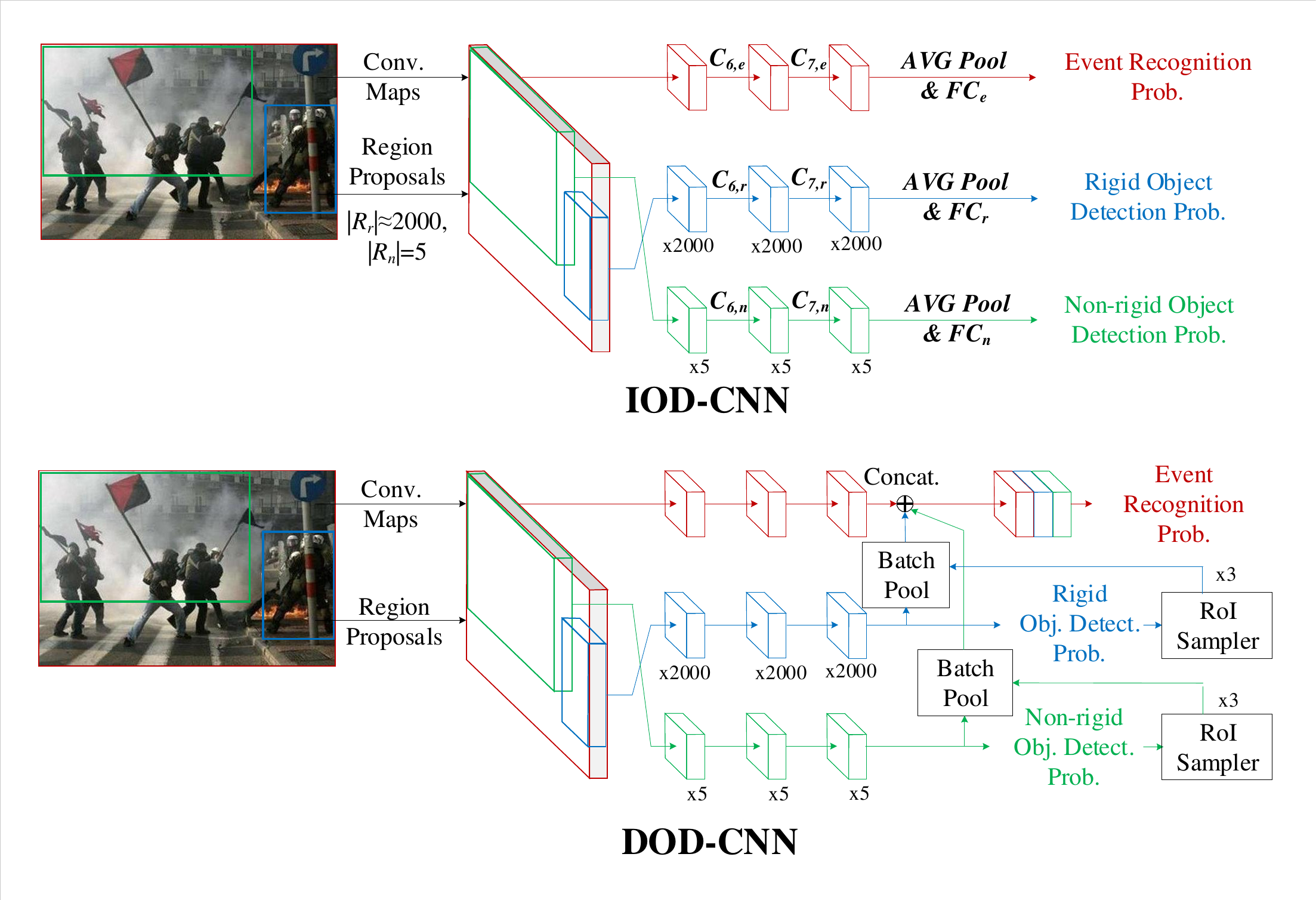}}
\end{minipage}
\vspace{-0.8cm}
\caption{\small{{\bf IOD-CNN and DOD-CNN Architectures.} \textcolor{myred}{Red}, \textcolor{myblue}{blue}, and \textcolor{mygreen}{green} arrows indicate the computational flow responsible for event recognition, rigid object detection, and non-rigid object detection, respectively.}}
\label{fig:architecture}
\end{figure}

In this paper, we adopt a function to directly inject the object detection output into the event recognition module to complement the IOD-CNN architecture which only utilizes the object information indirectly. We refer to our novel architecture as DOD-CNN (Doubly-injected Object Detection CNN). In this architecture, rigid object detection and non-rigid object detection network output feature maps are extracted from the intermediate layers for multiple regions-of-interests (RoIs). It then selects a certain number of feature maps with highest classification probabilities ($RoI$ $Sampler$) and aggregates them into one single map by using the {\it batch pooling}. Feature maps for rigid object detection and non-rigid object detection are then concatenated with the event recognition feature map at the same layer depth (after $C_7$ layers), where the concatenation is performed across the channel direction. The architectural comparison between the DOD-CNN and the IOD-CNN is illustrated in Figure~\ref{fig:architecture}.

We evaluated the proposed approach on the Malicious Crowd Dataset~\cite{HLeeICASSP2018}. The experiments demonstrate that utilizing the object detection information in both direct (injecting the feature maps) and indirect (transferring the information via shared weights) ways are effective in enhancing malicious event recognition performance.

Our contributions can be summarized as:
\begin{enumerate}
    \item Develop a novel architecture, DOD-CNN, which directly injects the object detection output into the event recognition network.
    \item Introduce a batch pooling layer which aggregates multiple feature maps corresponding to multiple RoIs into a single representative feature map.
    \item Demonstrate the effectiveness of DOD-CNN on the Malicious Crowd Dataset.
\end{enumerate}

\section{IOD-CNN}

IOD-CNN~\cite{SEumICIP2017} consists of shared layers, a RoI pooling layer, and three separate modules each responsible for event recognition, rigid object detection, and non-rigid object detection, respectively, as shown in Figure~\ref{fig:architecture}. Original architecture uses three fully connected (FC) layers (known as $FC_6$, $FC_7$, and $FC_8$) for each module. To adapt this architecture for DOD-CNN while considering the memory issues, the three FC layers in each module are replaced by two convolutional layers ($C_6$ and $C_7$), one average pooling layer, and one FC layer, where the output dimension of the FC layer is set to match the number of events or objects. Memory requirement significantly increases as the width of the last FC layer has to match the enlarged concatenated feature maps, and thus, the first and the second FC layers were pulled out. However, this modification cannot avoid a performance loss (from 93.6\% to 90.7\% in Table~\ref{tab:performance}) because newly adopted convolutional layers is learned from scratch instead of finetuning from the FC layers of any predefined network trained over a large-scale dataset.

IOD-CNN receives an input image and passes it through the shared layers. The RoI pooling layer takes in the per-image feature map which is the output of the shared layers along with three sets of RoIs generated for three tasks. While an entire region of the input image is used as the RoI for event recognition, selective search~\cite{JUijlingsIJCV2013} and multi-scale sliding windows~\cite{OViolaCVPR2001,NDalalCVPR2005,PFelzenszwalbTPAMI2010,HLeeACCV2012} are used to generate the RoIs for rigid and non-rigid object detection, respectively. For the rigid object detection, approximately 2000 RoIs are generated for each image while five RoIs are considered for the non-rigid object detection. For each RoI, per-RoI feature map is computed via RoI pooling and is fed into its corresponding task-specific module.

\section{DOD-CNN: Doubly-injecting Object Information}

\subsection{Architecture}

\begin{figure}[t]
\begin{minipage}[b]{1.0\linewidth}
  \centering
  \centerline{\includegraphics[width=\linewidth,trim=5mm 5mm 5mm 5mm,clip]{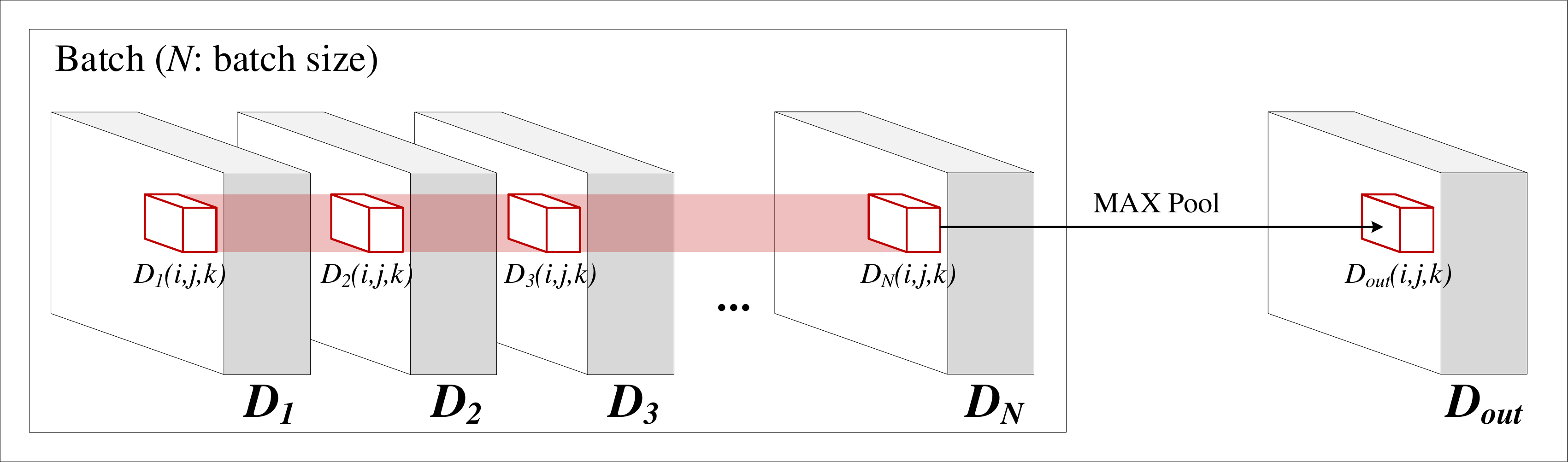}}
\end{minipage}
\vspace{-0.8cm}
\caption{\small{{\bf Batch Pooling Layer.} Batch pooling layer takes a batch with size $N$, and outputs a batch with size of one.}}
\label{fig:batch_maxpool}
\end{figure}

DOD-CNN adopts a novel function to pull out the intermediate output of object detection networks and directly inject them into event recognition networks, when compared to the baseline IOD-CNN architecture, as shown in Figure~\ref{fig:architecture}. Output feature maps from the $C_7$ layers in all three modules are concatenated across the channel direction. This concatenated map is pooled via an average pooling and then fed into the subsequent $FC_e$ layer of the event recognition module. If AlexNet~\cite{AKrizhevskyNIPS2012} is used as the backbone of our DOD-CNN, all maps share the same dimension as $6\times6\times256$ and the concatenated map becomes $6\times6\times768$. Accordingly, layers taking this concatenated map as an input should proportionally increase the filter depth, which then leads to a memory issue requiring an architectural modification when inheriting from IOD-CNN.

For each image, the network evaluates multiple RoIs for rigid and non-rigid object detections. Among multiple feature maps corresponding to these RoIs, we only use a selected number of RoIs with highest classification probabilities (three RoIs for our experiments). These selected feature maps are then transformed into a single map via the {\it batch pooling layer}. Note that the feature map selection process is performed to disregard the RoIs which are likely to contain no object information.\\

\noindent{\bf Batch Pooling Layer.} The function of this layer is to reduce the batch size as the output has to match the size of only one feature map. Batch pooling is computed as follows:
\begin{equation}
    D_{out}(i,j,k) = \max\{D_l(i,j,k)\}_{l=1,2,\cdots,N},
\end{equation}
where $D_l(i,j,k)$, $l=1,2,\cdots,N$ is an activation at a spatial location $(i,j)$, $k^{th}$ channel in $l^{th}$ feature map in a batch, where a batch contains $N$ feature maps. $D_{out}$ is the output feature map of the batch pooling layer which contains the most significant activations among all the feature maps in the batch. The process of the batch pooling is illustrated in Figure~\ref{fig:batch_maxpool}.

When learning DOD-CNN, the back-propagation does not pass through the batch pooling layer to hold back the event recognition network from affecting the other object detection network optimization.

\subsection{Training}

\begin{figure}[t]
\begin{minipage}[b]{1.0\linewidth}
  \centering
  \centerline{\includegraphics[width=\linewidth,trim=5mm 5mm 5mm 5mm,clip]{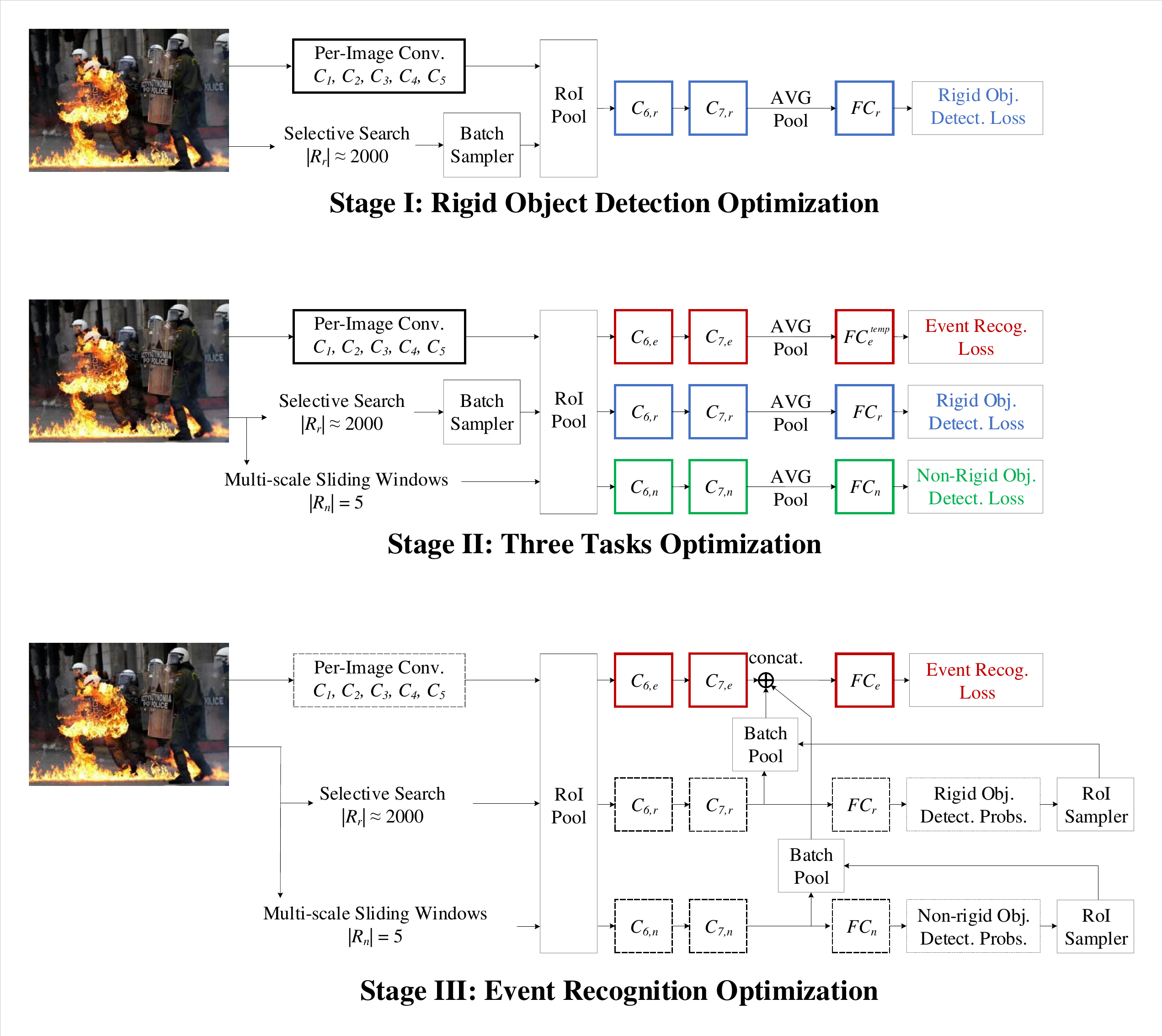}}
\end{minipage}
\vspace{-0.8cm}
\caption{\small{{\bf Three-stage Cascaded Optimization.} DOD-CNN is trained in three stages. {\bf Black}, \textcolor{myred}{red}, \textcolor{myblue}{blue}, and \textcolor{mygreen}{green} boxes indicate the shared layers, event recognition module, rigid object detection module, and non-rigid object detection module, respectively. Black dashed boxes indicate the layers which are not updated in the corresponding stage.}}
\label{fig:training}
\end{figure}

DOD-CNN is trained using a mini-batch stochastic gradient descent (SGD) optimization approach. A batch contains two images comprising one malicious image and one benign image. For event recognition and non-rigid object detection modules, 1 and 5 RoIs are generated per image, thus 2 and 10 RoIs are used as one batch, respectively. For training rigid object detection, approximately 2000 RoIs are generated for each image while a batch only takes 64 RoIs randomly selected from the entire RoI set of each image. Accordingly, a large number of batches are necessary to cover the entire RoI set for training rigid object detection.

To deal with this issue, we adopt a three-stage cascaded optimization strategy to train the DOD-CNN. In the first stage, only the layers used in performing rigid object detection are trained to allow more training iterations. Then, as the second stage, all three tasks are co-optimized in an end-to-end fashion in the second stage. In this stage, the event recognition network does not use the intermediate output from rigid and non-rigid object detections. This stage is used to learn the shared layers and two object detection modules. Note that the first and second stages are equivalently used to train the IOD-CNN in \cite{SEumICIP2017}. In the third stage, event recognition network is trained by directly injecting the object detection feature maps into the event recognition module while the weights in the shared layers and object detection modules are fixed. While the weights in $C_{6,e}$ and $C_{7,e}$ layers are inherited from the second stage, $FC_e$ layer is newly initialized according to a Gaussian distribution with mean of 0 and standard deviation of 0.01. $FC_e$ layer cannot be inherited from the previous stage because this layer has a different depth when compared with the corresponding layer ($FC_e^{temp}$ in Figure~\ref{fig:training}) in the previous stage. Figure~\ref{fig:training} illustrates the three stages used to train the DOD-CNN. For the first stage, we used the learning rate of 0.001, 50k iterations, and the step size of 30k. For the second and third stage, we trained with the learning rate of 0.0001, 20k iterations, and the step size of 12k.

To label the RoIs (for training purpose) in the rigid and non-rigid object detection, we have used 0.5 and 0.1 as the thresholds for the intersection over union (IoU) metric. We consider any RoI, whose IOU with respect to the ground truth bounding box is larger than the threshold, as a positive training example. For the rigid object detection, RoIs whose IoU is lower than 0.1 are used as negative training examples.

\subsection{Event Recognition}

While IOD-CNN only uses event recognition network for testing, DOD-CNN needs to utilize all the networks. This is because DOD-CNN directly utilizes the intermediate output and classification probabilities from the object detection networks. This may increase a memory size. Please note that replacing {\it FC} by {\it C} significantly reduced the required memory size, which will compensate the increased memory size caused by including the object detection networks for testing.


\section{Experiments}

\subsection{Dataset}

We have selected the Malicious Crowd Dataset~\cite{HLeeICASSP2018,SEumDCS2018} as it provides the appropriate components in evaluating the effect of using object information for event recognition. The dataset contains 1133 crowd images and is equally split into malicious and benign classes. For evaluation, half of each class is used for training and the remaining is used for testing. Along with the event class labels, bounding box annotations for three rigid objects ({\it police}, {\it helmet}, and {\it car}) and two non-rigid objects ({\it fire} and {\it smoke}) are provided. Details on how these objects are selected are given in \cite{HLeeICASSP2018}.

\subsection{Performance Evaluation}

\begin{table}[t]
\setlength{\tabcolsep}{20.5pt}
\renewcommand{\arraystretch}{1.1}
\begin{center}
\begin{tabular}{c|c|c}
\specialrule{.15em}{.05em}{.05em}
\multicolumn{2}{c|}{Method} & AP (\%) \\
\specialrule{.15em}{.05em}{.05em}
\multirow{2}{*}{IOD-CNN} & Original~\cite{SEumICIP2017} & 93.6 \\
& Modified & 90.7 \\\hline
\multicolumn{2}{c|}{DOD-CNN} & {\bf 94.6} \\
\specialrule{.15em}{.05em}{.05em}
\end{tabular}
\end{center}
\vspace{-0.6cm}
\caption{\small{{\bf Event recognition} average precision (AP).}}
\label{tab:performance}
\end{table}

To demonstrate the effectiveness of exploiting object detection output for event recognition, we compare the DOD-CNN with two baselines: IOD-CNN~\cite{SEumICIP2017} and its modified version which replaces three {\it FC} layers with two convolutional layers and one {\it FC} for each module. Their event recognition accuracies are shown in Table~\ref{tab:performance}. DOD-CNN outperforms the two baselines by at least 1.0\% AP which shows that it is a better practice to include (in addition to the indirect sharing) the direct injection of the intermediate object detection network outputs to the event recognition module. We also note that modified IOD-CNN architecture under-performs than the original architecture by 2.9\%. This performance drop may have been caused by using the layers which are not inherited from the network trained on a large-scale Places dataset~\cite{BZhouNIPS2014}.

\begin{table}[t]
\setlength{\tabcolsep}{11.0pt}
\renewcommand{\arraystretch}{1.1}
\begin{center}
\begin{tabular}{c|c|c|c}
\specialrule{.15em}{.05em}{.05em}
Task & Single-task & IOD-CNN & DOD-CNN \\
\specialrule{.15em}{.05em}{.05em}
E &  89.9 & 90.7 & {\bf 94.6} \\
R & {\bf 8.1} & 7.8 & 7.8 \\
N & 30.4 & {\bf 37.2} & {\bf 37.2} \\
\specialrule{.15em}{.05em}{.05em}
\end{tabular}
\end{center}
\vspace{-0.6cm}
\caption{\small{{\bf Single task versus multitask performance.} Task: {\bf E}: Event Recognition, {\bf R}: Rigid Object Detection, {\bf N}: Non-rigid Object Detection.}}
\label{tab:sigle_vs_multi}
\end{table}

We have also carried out an experiment to analyze how the performance of each task changes when applying IOD-CNN (indirect sharing) or DOD-CNN (both indirect and direct injection of object information). Note that the optimization of DOD-CNN and IOD-CNN share the first and second stages in the training process while DOD-CNN takes an additional third stage where only the event recognition module is trained with directly injected object detection output features. Therefore, task-wise performances for rigid and non-rigid object detection modules in the IOD-CNN and DOD-CNN are naturally equivalent. Table~\ref{tab:sigle_vs_multi} shows that, when the tasks are co-learned, non-rigid object detection performance is boosted while rigid object detection performance is sacrificed.

\subsection{Ablation Studies}

\noindent{\bf Injection Location of Object Detection Output.} To find the most effective way to directly inject the object detection output into the event recognition module via feature map concatenation, we have conducted an experiment to explore three different concatenation locations in the network. Table~\ref{tab:concatlayer} shows that the best location to adopt the feature map concatenation is after $C_7$, which is closer to the end of the network.\\

\begin{table}[t]
\setlength{\tabcolsep}{3pt}
\renewcommand{\arraystretch}{1.1}
\begin{center}
\begin{tabular}{c|c|ccc}
\specialrule{.15em}{.05em}{.05em}
& \multirow{2}{*}{IOD-CNN} & \multicolumn{3}{c}{DOD-CNN: Direct Injection Location} \\\cline{3-5}
& & ~~~~~~$C_6$~~~~~~ & ~~~~~~$C_7$~~~~~~ & $C_6$ \& $C_7$ \\\specialrule{.15em}{.05em}{.05em}
AP (\%) & 90.7 & 91.4 & {\bf 94.6} & 93.2 \\
\specialrule{.15em}{.05em}{.05em}
\end{tabular}
\end{center}
\vspace{-0.6cm}
\caption{\small{{\bf Performance comparison w.r.t. the injection location of object detection output.} Note that IOD-CNN does not adopt the direct injection of object detection output.}}
\label{tab:concatlayer}
\end{table}

\noindent{\bf Effect of Late Fusion.} We have also applied a late fusion approach over the final scores of all three tasks. We have selected the dynamic belief fusion (DBF) \cite{HLeeWACV2016} as the fusion method. Table~\ref{tab:latefusion} presents the event recognition performance of IOD-CNN and DOD-CNN after applying the late fusion. The performance of the IOD-CNN was enhanced via a late fusion by 0.6\% AP, while only 0.1\% AP has been gained by the combination of DOD-CNN and DBF. This observation suggests that the DOD-CNN contains the functionality of the late fusion embedded into a network itself, as it directly integrates the intermediate outputs of the three tasks. Thus, practically, DOD-CNN requires no additional late fusion approaches at the end of the network for performance boost.\\

\begin{table}[t]
\setlength{\tabcolsep}{12.5pt}
\renewcommand{\arraystretch}{1.1}
\begin{center}
\begin{tabular}{c|cc|c}
\specialrule{.15em}{.05em}{.05em}
Method & w/o DBF & w/ DBF & gain \\\specialrule{.15em}{.05em}{.05em}
IOD-CNN~\cite{SEumICIP2017} & 93.6 & 94.2 & +0.6 \\
DOD-CNN & 94.6 & 94.7 & +0.1 \\
\specialrule{.15em}{.05em}{.05em}
\end{tabular}
\end{center}
\vspace{-0.6cm}
\caption{\small{{\bf Late fusion performance (DBF~\cite{HLeeWACV2016}).}}}
\label{tab:latefusion}
\end{table}

\section{Conclusion}

We have developed a CNN-based event recognition approach which is referred to as DOD-CNN. This architecture is inspired by the architecture of IOD-CNN, where object detection information is indirectly exploited for enhancing the event recognition performance by utilizing the shared layers between those two. In this architectural integration, our network also adopts the function of using object detection output for event recognition into the architecture. The experimental results show that taking advantage of both indirect and direct way of injecting the object information for the event recognition within DOD-CNN is highly effective in boosting the performance.

\bibliographystyle{IEEEbib}
\bibliography{refs}

\end{document}